\documentclass[10pt,twocolumn,letterpaper]{article}

\usepackage{iccv}
\usepackage{times}
\usepackage{epsfig}
\usepackage{graphicx}
\usepackage{amsmath}
\usepackage{amssymb}
\usepackage{algorithm}
\usepackage{algorithmic}
\usepackage{bbm}
\usepackage{amsfonts}
\usepackage{mathrsfs}
\usepackage{multirow}
\usepackage{subfig}
\usepackage{booktabs,siunitx}
\usepackage[pagebackref=true,breaklinks=true,letterpaper=true,colorlinks,bookmarks=false]{hyperref}

\iccvfinalcopy % *** Uncomment this line for the final submission

\def\iccvPaperID{406} % *** Enter the CVPR Paper ID here

\usepackage{siunitx}
\usepackage{enumitem}

\newcommand{\ModelName}{MiVOLO}
\newcommand{\FullDatasetName}{Layer Age Gender}
\newcommand{\DatasetName}{Layer Age and Gender Dataset}
\newcommand{\DatasetNameShort}{LAGENDA}

\begin{document}

%%%%%%%%% TITLE
\title{\ModelName: Multi-input Transformer for Age and Gender Estimation}

\author{
\begin{tabular}[t]{c@{\extracolsep{8em}}c} 
Maksim Kuprashevich  & Irina Tolstykh \\
mvkuprashevich@gmail.com & irinakr4snova@gmail.com
\end{tabular}\\
\\
Layer Team, SaluteDevices
}

\maketitle
\thispagestyle{empty}

\begin{abstract}Age and gender recognition in the wild is a highly challenging task: apart from the variability of conditions, pose complexities, and varying image quality, there are cases where the face is partially or completely occluded. We present \ModelName \ (Multi Input VOLO), a straightforward approach for age and gender estimation using the latest vision transformer. Our method integrates both tasks into a unified dual input/output model, leveraging not only facial information but also person image data. This improves the generalization ability of our model and enables it to deliver satisfactory results even when the face is not visible in the image. To evaluate our proposed model, we conduct experiments on five popular benchmarks and achieve state-of-the-art performance, while demonstrating real-time processing capabilities. Additionally, we introduce a novel benchmark based on images from the Open Images Dataset. The ground truth annotations for this benchmark have been meticulously generated by human annotators, resulting in high accuracy answers due to the smart aggregation of votes. Furthermore, we compare our model's age recognition performance with human-level accuracy and demonstrate that it significantly outperforms humans across a majority of age ranges.
Finally, we grant public access to our models, along with the code for validation and inference. In addition, we provide extra annotations for used datasets and introduce our new benchmark. The source code and data can be accessed at \url{https://github.com/WildChlamydia/MiVOLO.git}
\end{abstract}

\section{Introduction}

Age and gender recognition of a person in a photo is a highly important and complex task in computer vision. It is crucial for various real-world applications, including retail and clothes recognition, surveillance cameras, person identification, shopping stores and more. Additionally, this task becomes even more challenging in uncontrolled scenarios. The significant variability of all conditions such as image quality, angles and rotations of the face, partial facial occlusion, or even its absence in the image, coupled with the necessary speed and accuracy in real-world applications, makes the task quite challenging.

Our objective was to develop a simple and easy to implement approach capable of simultaneously recognizing both age and gender, even in situations where the face is not visible. We aimed for scalability and speed in our solution.

In this paper, "gender recognition" refers to a well-established computer vision problem, specifically the estimation of biological sex from a photo using binary classification. We acknowledge the complexity of gender identification and related issues, which cannot be resolved through a single photo analysis. We do not want to cause any harm to anyone or offend in any way.

\begin{figure}[t]
\centering
\includegraphics[width=8.2cm]{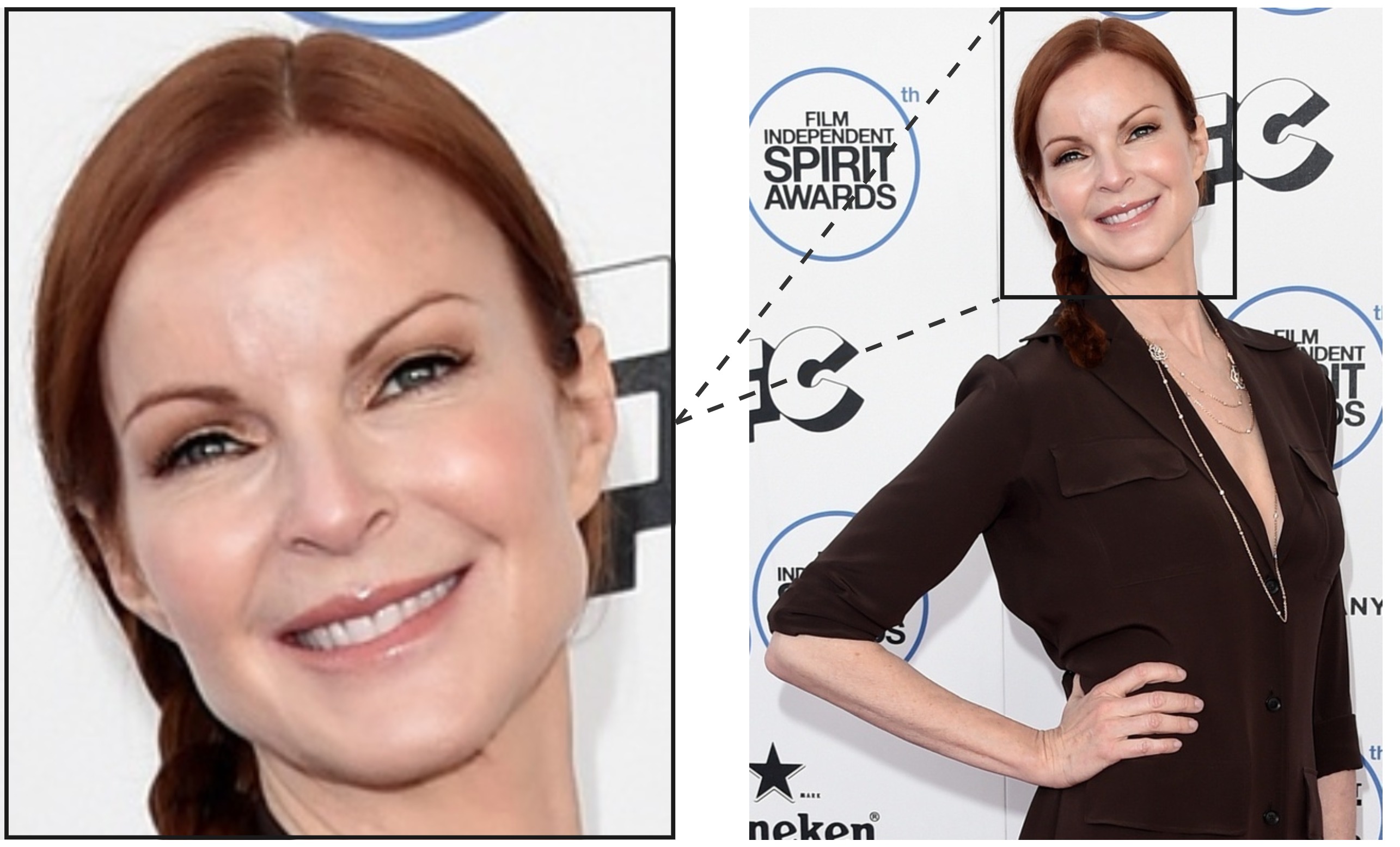}\\
\caption{Illustration of the complexity of the age recognition task. The image from \cite{imdbwikiclean} features actress Marcia Cross at the 30th Film Independent Spirit Awards in 2015 \cite{marcia_at_reward}. Marcia was born in 1962, making her 53 years old in the picture. The average error for this image during crowd source annotation (used as a honeypot) was 16.17 years (n=43). Due to Marcia's youthful appearance in this photo, users were prone to making significant estimation errors.}\label{fig:1}
\end{figure}

Meanwhile, gender recognition is a classification task, while age estimation can be solved either through regression or classification.

Many popular benchmarks and research papers \cite{adience} \cite{fairface} consider age as a classification problem with age ranges. However, this approach can be inaccurate because age, by its nature, is a regression problem. Moreover, it is inherently imbalanced \cite{dir}. Treating it as a classification causes several issues. For a classification model, it makes no difference whether it misclassifies into a neighboring class or deviates by several decades from the ground truth. Additionally, as stated in \cite{dir}, classification models cannot approximate age ranges from unseen classes, while regression models can. However, regression models are much trickier to train, and collecting or cleaning datasets for such task is more challenging.

In this paper, we consider the following popular benchmarks: IMDB-Clean \cite{imdbwikiclean}, UTKFace \cite{utk}, Adience \cite{adience}, FairFace \cite{fairface}, AgeDB \cite{moschoglou2017agedb}. These are some of the most famous datasets containing both age and gender ground truth. IMDB-Clean is the largest available dataset for this task, but it consists of celebrities and is heavily biased. This bias poses a problem for recognition in the wild, you can see an example in Figure \ref{fig:1}. For more details, refer to the \ref{section:datasets} section.
Therefore, in our work, we introduce a completely new \FullDatasetName\ benchmark comprising 84,192 pairs $(FaceCrop, BodyCrop)$ randomly selected from the Open Images Dataset\cite{oid4}.
These images were annotated on a crowd-sourcing platform, and we have achieved remarkably high accuracy using a weighted averaging votes strategy. 

While most existing works focus on estimating age and/or gender solely from face images, this work introduces the MiVOLO model, which is built upon the visual transformer model VOLO \cite{volo}. The \ModelName\ allows for the simultaneous prediction of age and gender by incorporating both face and body features.

Our model, trained using both body and face images, achieves SOTA results on the 5 largest benchmarks.
Additionally, it attains a high frame rate of 971 frames per second (FPS) when utilizing a batch size of 512 on the NVIDIA V100. Moreover, our model accommodates the inclusion of images that may lack visible faces.

Human-level estimation is also an open question. Accuracy heavily depends on conditions and is unclear in this task. Some articles \cite{humanlevel1} state that neural network models have already surpassed human-level performance. However, there are not many works where exact human-level performance has been estimated, and we did not find any that have been conducted on images with full-sized persons in the wild.
In this paper, we estimated this level using random images from the IMDB-clean dataset.

The main contributions of our work can be summarized as follows:
\begin{itemize}
\vspace{-0.2cm}\item We provide publicly available models that achieved SOTA results in 5 benchmarks.
\vspace{-0.2cm}\item We have developed a readily implementable architecture called \ModelName, capable of simultaneously handling faces and bodies. It enables accurate age and gender prediction, even in cases where humans may struggle. The architecture supports predictions with and without face input. \ModelName \ has achieved top-1 results on 5 popular benchmarks, 3 of them without any fine-tuning on training data.
\vspace{-0.2cm}\item Additionally, we have once again demonstrated that a carefully implemented multi-output (multi-task) approach can provide a significant performance boost compared to single-task models.
\vspace{-0.2cm}\item We have also shown that multi-input models able to gain generalization ability in the same way as multi-task models.
\vspace{-0.2cm}\item The original UTKFace dataset has been restored to include full-sized images.
\vspace{-0.2cm}\item The annotations of IMDB-clean, UTK and FairFace datasets have been modified to include annotations of all detectable persons and faces in each image using our models.
\vspace{-0.2cm}\item Human-level estimation for the task with a substantial sample size.
\vspace{-0.2cm}\item A completely new, very well balanced \DatasetName \ that we propose to use as a benchmark for age and gender recognition in the wild.
\end{itemize}
\section{Related Works} \label{section:related_work}
\noindent\textbf{Facial age and gender recognition.} Typically, solving the gender task separately is not of great interest in research or business. Therefore, most methods and benchmarks either tackle both age and gender tasks or focus solely on age. Convolutional neural networks (CNNs) have become the state-of-the-art in most computer vision challenges, although in recent years, there has been a trend to replace them in certain tasks. Levi et al. \cite{levi} were the first to use CNNs, evaluating their approach on the Adience dataset \cite{adience}, which contains age and gender as classes. 

The network they implemented is a convolutional model with two fully-connected layers. It achieves an accuracy of 50.7 ± 5.1 for age. 
With significant advancements in computer vision neural networks, many methods have been developed, some based on face recognition techniques and models \cite{retina_arc}, suggesting the existence of powerful generic models for faces that can be adapted to downstream tasks.
Some papers \cite{ordinal_regress} even employ more general models as encoders, such as VGG16 \cite{vgg}, particularly for ordinal regression approaches in age estimation.	
Other methods utilize CNN networks for direct classification or regression for age recognition \cite{reg_convnets} \cite{deep_ordinal}.	
As of the writing of this article, the state-of-the-art model on Adience for age classification used the Attention LSTM Networks approach \cite{lstm_age}, achieving an accuracy of 67.47. However, they did not employ their model for gender prediction.

\noindent\textbf{Recognition using face and body images.} 
Most methods for age or gender estimation are based on facial analysis. Some consider the body for age \cite{body_survey} or gender \cite{human_body_gender1} recognition, but in very few works \cite{bonet_face_person}, joint recognition using both face and body pictures has been utilized. Therefore, it is difficult to find a baseline in open sources to start with.
Only a few works exist that utilize full-body images of individuals. The earliest attempt \cite{age_body_no_network} predates the era of neural networks and employed classical image processing techniques to predict age.
A more recent study \cite{bonet_face_person} utilized both face and body images together in a single neural network for age and gender prediction. Another paper \cite{body_gender} employed face and body images with a late fusion approach, but solely for gender prediction.

\noindent\textbf{Datasets and benchmarks.} Our focus primarily lies on datasets containing both age and gender information. The largest existing dataset for these tasks is IMDB-Wiki \cite{imdb-wiki-orig1} \cite{imdb-wiki-orig2}. However, the ground truth answers in this dataset do not appear to be clean. Therefore, we used the cleaned version  \cite{imdbwikiclean}.	
Another interesting dataset is UTKFace \cite{utk}, which also contains both age and gender information but is much smaller, with only annotations for face crops.	
The MORPH \cite{morph} dataset is also notable for age estimation, although the domain captured in this dataset cannot be considered as representing wild conditions.	
KANFace \cite{kanface} is another large dataset of images and videos that includes gender information.	
The CACD dataset \cite{cacd} is also sizeable and features celebrities from different age groups, making it highly useful, but it does not include gender information.
The AgeDB \cite{moschoglou2017agedb} dataset contains face images of celebrities with age variations ranging from 1 to 101 years old, encompassing two gender groups.
The aforementioned datasets above contains age information suitable for regression. \\
Adience dataset \cite{adience} contains both age and gender, but age presented as 8 classes.
FairFace \cite{fairface} is a big and well-balanced dataset, where age is categorized into ranges.
All these datasets are focused on faces, but for most of it is possible to generate additionally persons information.
We are using for training experiments only IMDB-clean and UTKFace as biggest datasets with suitable image domain and annotations.
The FairFace, Adience and AgeDB are employed specifically for benchmarking purposes.

\noindent\textbf{Visual Transformer Models.} For many years, convolutional neural networks have dominated the field of computer vision. However, transformers have been increasingly gaining prominence in various tasks and benchmarks. Transformers are powerful and versatile models, and they are far from reaching their limits. One of the first transformer models applied to computer vision was ViT \cite{vit}, which achieved great success and inspired the exploration of many other variants \cite{xcit} \cite{cait}. VOLO\cite{volo} is also a transformer-based model, but it efficiently combines the worlds of CNNs and Transformers and performs extremely well. We chose the VOLO model because it converges quickly and requires less data in our experience. Additionally, VOLO is one of the fastest transformer-based vision models.

\noindent\textbf{Human level for age estimation.} 
In \cite{human_vs_machine}, a comparison was made between the mean absolute error (MAE) of human and machine age estimation. The study examined the FG-NET dataset and the Pinellas County Sheriff's Office (PCSO) dataset (currently unavailable). The authors found that the human MAE on the FG-NET dataset \cite{fgnet} was 4.7, while on the PCSO dataset it was 7.2. For the machine results, they obtained 4.6 and 5.1, respectively. They also claimed that their algorithm performed worse than humans in the age range $\in[0, 15]$ years. The authors noted that this age range is not present in the FG-NET dataset \cite{fgnet}, which caused the observed difference. When excluding this range, the estimated human MAE for FG-NET is also very close - 7.4. Eventually, the authors concluded that their model is more accurate than humans. \\
\section{Datasets} \label{section:datasets}

\subsection{IMDB-clean}

\begin{figure}[t]
\centering
\includegraphics[width=8.2cm]{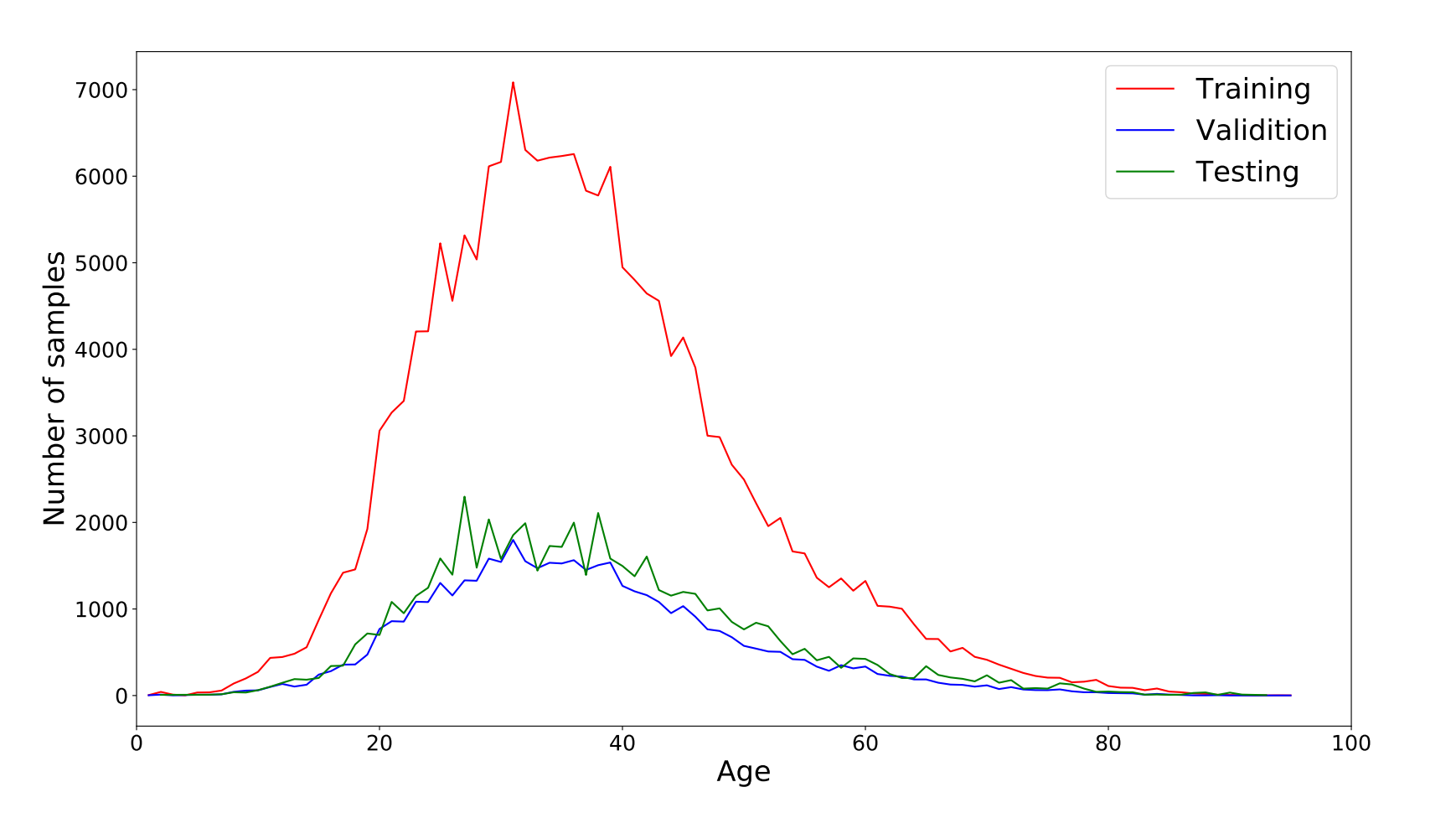}\\
\caption{Age distributions of the IMDB-clean from original paper \cite{imdbwikiclean}.}
\label{fig:imdb_distr} 
\end{figure}

We primarily conducted our experiments on the IMDB-Clean dataset, which comprises 183,886 training images, 45,971 validation images, and 56,086 test images. We utilized the original split of the dataset. The images in this dataset are highly suitable for our tasks and represent a wild domain. However, it is important to note that the dataset only includes images of celebrities, which introduces a bias. Additionally, it suffers from significant class imbalance (Figure \ref{fig:imdb_distr}), similar to other datasets.

For the face-only baseline, we utilized this dataset without making any modifications. For experiments involving body images, we generated face and person bounding boxes for all individuals detectable with our model in each image.

\subsection{UTKFace}

The original dataset only includes bounding boxes for cropped face images, as we performed backward matching to the original full-sized images. This process also involved double-checking by the face encoder. During this process, we encountered 4 faces that did not match back to the original images, so we dropped those images from our annotation. The remaining images maintain the original annotations but with bounding boxes generated by our detector.

The original dataset does not provide any predefined training, validation, and test splits. To align with other works, we utilized the exact same split as described in \cite{utk_split}. In this subset, the ages are in range $\in [21, 60]$, totalling in 13,144 training and 3,287 test images.

\subsection{FairFace}

The FairFace\cite{fairface} dataset comprises 86,744 training images and 10,954 validation images. The ground truth attributes in this dataset cover race, gender, and age, categorized into nine classes: (0-2, 3-9, 10-19, 20-29, 30-39, 40-49, 50-59, 60-69, 70+). The dataset is also very well balanced by races.

For measuring gender and age classification accuracy, we utilize a validation set.

To gather information about bodies, we utilize 'fairface-img-margin125' images and employ our detector model to localize the centered face and its corresponding body region.

\subsection{Adience}

The Adience\cite{adience} dataset consists of 26,580 facial images, depicting 2,284 subjects across eight age group classes (0-2, 4-6, 8-13, 15-20, 25-32, 38-43, 48-53, 60-100). The dataset additionally includes labels for the binary gender classification task.

The images in the Adience\cite{adience} dataset were captured under varying conditions, including differences in head pose, lighting conditions, and image quality.

For our analysis, we utilize coarse aligned images and refrain from applying any other aligning methods. 
To refine the facial localization on the images, we employ our detector model. We deliberately avoid using in-plane aligned versions of the faces to prevent distortion. The validation of our models and computation of classification accuracy are performed using all five-fold sets.

\subsection{AgeDB}

The AgeDB\cite{moschoglou2017agedb} dataset includes 16,488 facial images. The annotations per image include gender and age.
Since each image in the dataset is a face crop, we take center crop and use it as input for our models. We use all images for validation purpose.

\subsection {New \DatasetName}

\subsubsection{\DatasetNameShort \ benchmark}

Due to issues such as bias in datasets containing celebrities and professional photos, we introduce a completely new benchmark in our paper for age and gender recognition tasks in wild conditions. We named this benchmark the \FullDatasetName \ Dataset (\DatasetNameShort), by the name of our team. To create it, we initially sampled random person images from the Open Images Dataset \cite{oid4} (OID). This dataset offers a high level of diversity, encompassing various scenes and domains.

The images were annotated using a crowd source platform. To ensure high-quality annotations for age estimation, we implemented strict control measures. Control tasks (honeypots) were included to maintain accuracy. Each honeypot had a 7-year age range, within $\pm3$ years of the true age. Therefore, the accuracy on these control tasks can be seen as just CS@3 (see \ref{section:metrics}).

Control measures included:

\begin{itemize}
\vspace{-0.2cm}\item Mandatory training for all users before proceeding.
\vspace{-0.2cm}\item Users had to pass an examination; CS@3 below 20\% resulted in a ban.
\vspace{-0.2cm}\item Annotation tasks consisted of 6 examples and 1 hidden control task, totaling 7 tasks per suite.
\vspace{-0.2cm}\item After completing 10 task suites, users with average CS@3 below 20\% were banned, and their answers rejected.
\end{itemize}

These measures were implemented to prevent significant noise from bots and cheaters.

Our dataset was annotated with an overlap of 10, meaning that each real task received 10 votes for both age and gender.

\begin{figure}[t]
\centering
\includegraphics[width=8.2cm]{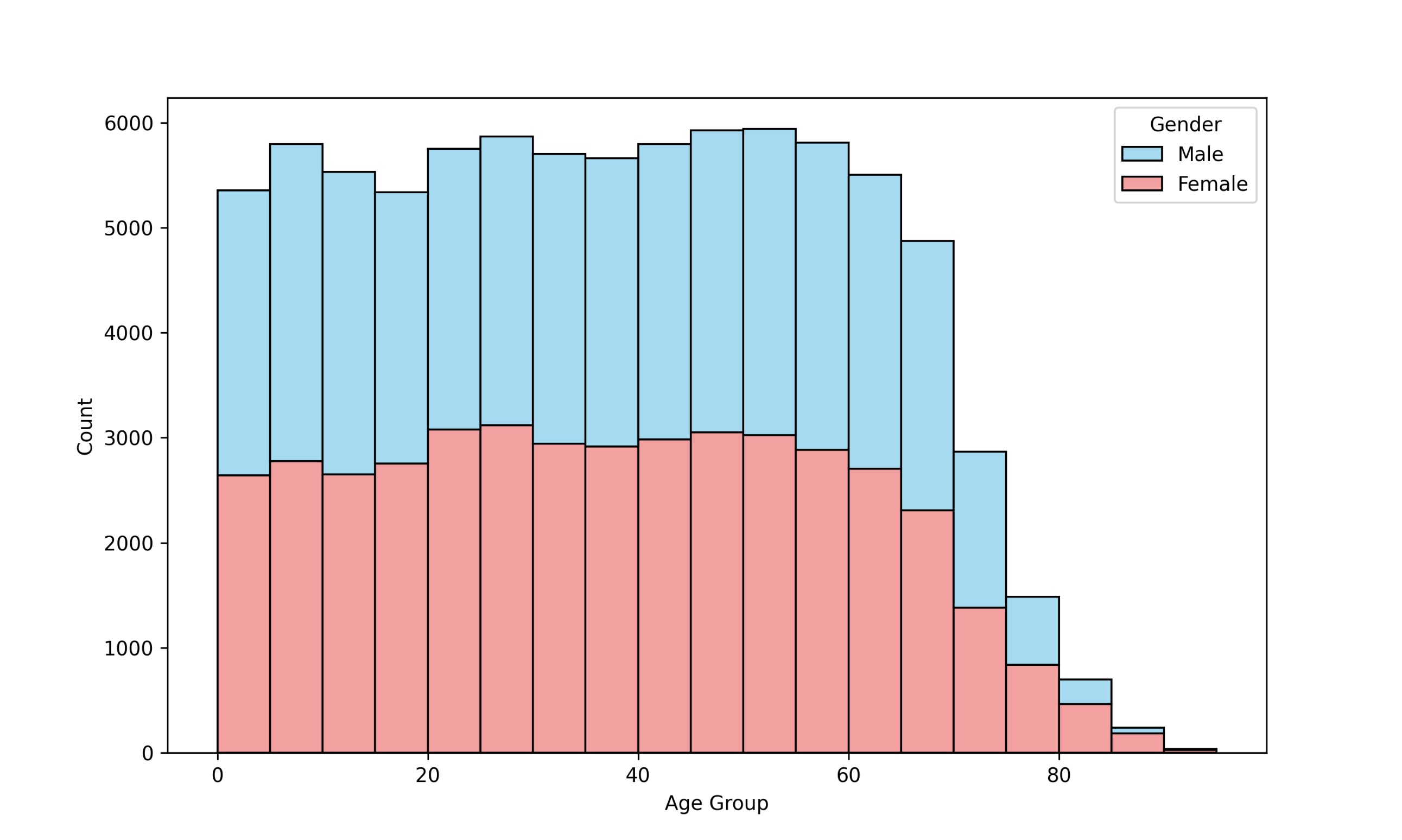}\\
\caption{Age and gender distributions with bin steps of 5 in the \DatasetNameShort \ result in an almost perfectly balanced benchmark. However, there are still imbalances in some ranges on the far right due to the difficulty in finding samples for these ages.}
\label{fig:data_distr} 
\end{figure}

In the last step, we balanced the dataset by age distribution using 5-year groups and ensured gender distribution within each one. As a result, we obtained 67,159 images with 84,192 persons, comprising 41,457 males and 42,735 females samples. Please refer to Figure \ref{fig:data_distr} for a visualization of the dataset distribution.

\subsubsection{Votes ensembling} \label{subsection:votes}

After completing the annotation process, we encountered the challenge of determining how to utilize the obtained votes. 

Table \ref{table:votes_aggregation} provides a list of all the methods that were tested. In addition to other statistical methods, we employed a weighted mean strategy. It was implemented as follows:
\[A(v) = \frac{\sum_{i=1}^{N} v_{i} * e^{(MAE(u_{i}))^{-1}}}{\sum_{i=1}^{N}e^{(MAE(u_{i}))^{-1}}} \]
where ~$A$~ is final age prediction for the $v$ vector of user votes, ~$N$~ is size of $v$, amount of users who annotated this sample and ~$MAE(u_{i})$~ denotes the individual MAE across all control tasks for the $i$-th user $u$. \\
We used an exponential weighting factor because there is a substantial difference in annotation quality between users with MAE of 3 and 4, for example. This approach outperformed other variants significantly.\\

Gender was aggregated using the simple $mode(v)$, where $v$ is an array of elements $\in{ male, female }$. We discarded all answers where the mode occurred with a frequency of less than 75\%. Based on control tasks, the gender accuracy has to be 99.72\%. We can roughly claim that human accuracy for this task is less or equal to this level.

\begin{table}[h!]
\centering
\begin{tabular}{|l|c|c|} 
 \hline
 \multicolumn{1}{|c|}{Method} & \multicolumn{1}{|c|}{MAE} & \multicolumn{1}{|c|}{CS@5, \%} \\ [0.5ex] 
 \hline\hline
 Mean & 4.77 & 62.43 \\ 
 Median & 4.75 & 65.44 \\ 
 Interquartile Mean & 4.74 & 63.80 \\
 Mode & 5.70 & 59.28 \\
 Maximum Likelihood & 4.81 & 65.86 \\
 Winsorized mean (6) & 4.73 & 63.44 \\
 Truncated mean (0.3) & 4.75 & 63.62 \\
 \textbf{Weighted mean} & \textbf{3.47} & \textbf{74.31} \\[1ex] 
 \hline
\end{tabular}
\caption{Different statistic methods to aggregate N votes into one age prediction.}
\label{table:votes_aggregation}
\end{table}

\subsubsection{\DatasetNameShort \ trainset}

Experiments in this work required not only a high-quality benchmark but also a large amount of training data. Therefore, besides our benchmark, we also collected data from other sources, mostly from our production. These images are in almost the same visual domain as images from OID\cite{oid4}. 

Our train dataset contains approximately 500,000 images in total, which have been annotated in exactly the same way as \DatasetNameShort\ benchmark.

In the text, we refer to this training and validation proprietary data as \DatasetNameShort \ trainset.
Although we cannot make this data publicly available, we provide a demo with the model trained on it (the link can be found in the Github repository). 
\section{Method}
\label{section:method}

\subsection{\ModelName: Multi-input age \& gender model } \label{section:multi_input_data_prep}

\begin{figure*}
  \includegraphics[width=\textwidth]{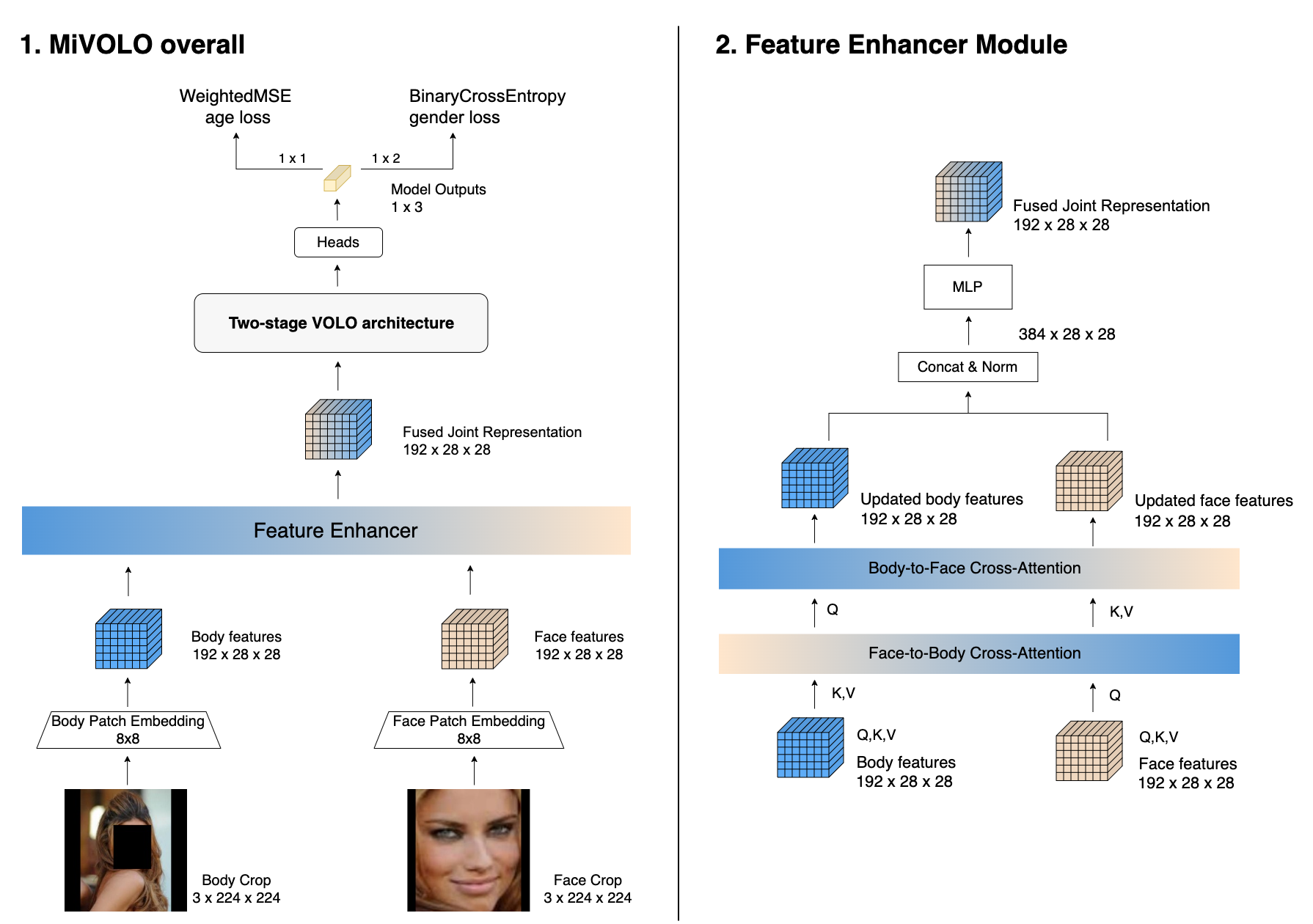}
  \caption{\ModelName. We present the overall model and a feature enhancer module in block 1 and block 2 respectively.}
  \label{fig:mivolo}
\end{figure*}

Our model is depicted in Figure \ref{fig:mivolo}.

For each input pair $(FaceCrop, BodyCrop)$ of size 224 × 224, we independently apply the original VOLO\cite{volo} patch embedding module, which tokenizes the crops into image patches of size 8 × 8.

Two representations are then fed into a feature enhancer module for cross-view feature fusion, which is achieved using cross-attention. The module is illustrated in Fig.~\ref{fig:mivolo}, block 2. Once the features are enriched with additional information, we perform a simple concatenation, followed by a Multi-Layer Perceptron (MLP) that creates a new fused joint representation and reduces the dimensionality of the features. 

This feature fusion allows us to pay attention to important features from both inputs and disregard less significant ones. Additionally, it handles scenarios where one of the inputs is empty, ensuring meaningful information is extracted even from a single view.

The fused features are then processed using the VOLO two-stage architecture\cite{volo}, which involves a stack of Outlookers, tokens downsampling, a sequence of transformers, and two heads on top.

The last two linear layers update the class embedding into a 3-dimensional vector: two output values for gender and one for the normalized age value. Unlike \cite{age_gender_multi_task}, which uses multiple heads for separate age and gender predictions, \ModelName\ produces a single vector for each image containing both outputs.

We use combination of two losses for training:

\begin{itemize}
\vspace{-0.2cm}\item WeightedMSE loss function for age prediction with weights from LDS\cite{dir}
\vspace{-0.2cm}\item BinaryCrossEntropy loss function for gender prediction
\end{itemize}

As demonstrated in Table \ref{table:baseline_age_gender_results}, multi-task learning enables us to achieve improvements in both tasks.

Moreover, early feature fusion allows us to maintain almost the same high performance as that of the original VOLO (see \ref{section:performance}).

\subsection{Data preprocessing} \label{section:multi_input_data_prep}

Each face and body crop image we resize by using letterbox with padding to preserve the aspect ratio, followed by RGB channel Z-score normalization, using the Imagenet original values. The resize algorithm used is bilinear. 

The ground truth answers are also processed with min-max normalization:
$$ \tilde{y}_i = \frac{y_i - y_{\min}}{y_{\max} - y_{\min}}. $$

To obtain face-body pairs, we follow these steps:
\begin{enumerate}
\vspace{-0.2cm}\item The $input$ image is first passed through a detector to find all faces and persons. We specifically trained YOLOv8 \cite{yolov8} for the publicly available version of our code.
\vspace{-0.2cm}\item Using the lists of face and person objects obtained, we run the $Assign(faces, persons)$ algorithm to associate faces with corresponding persons. This method makes use of the Hungarian algorithm. Unassigned faces or bodies they can still be utilized as independent inputs.
\end{enumerate}

\begin{figure}[t]
\centering
\includegraphics[width=7.5cm]{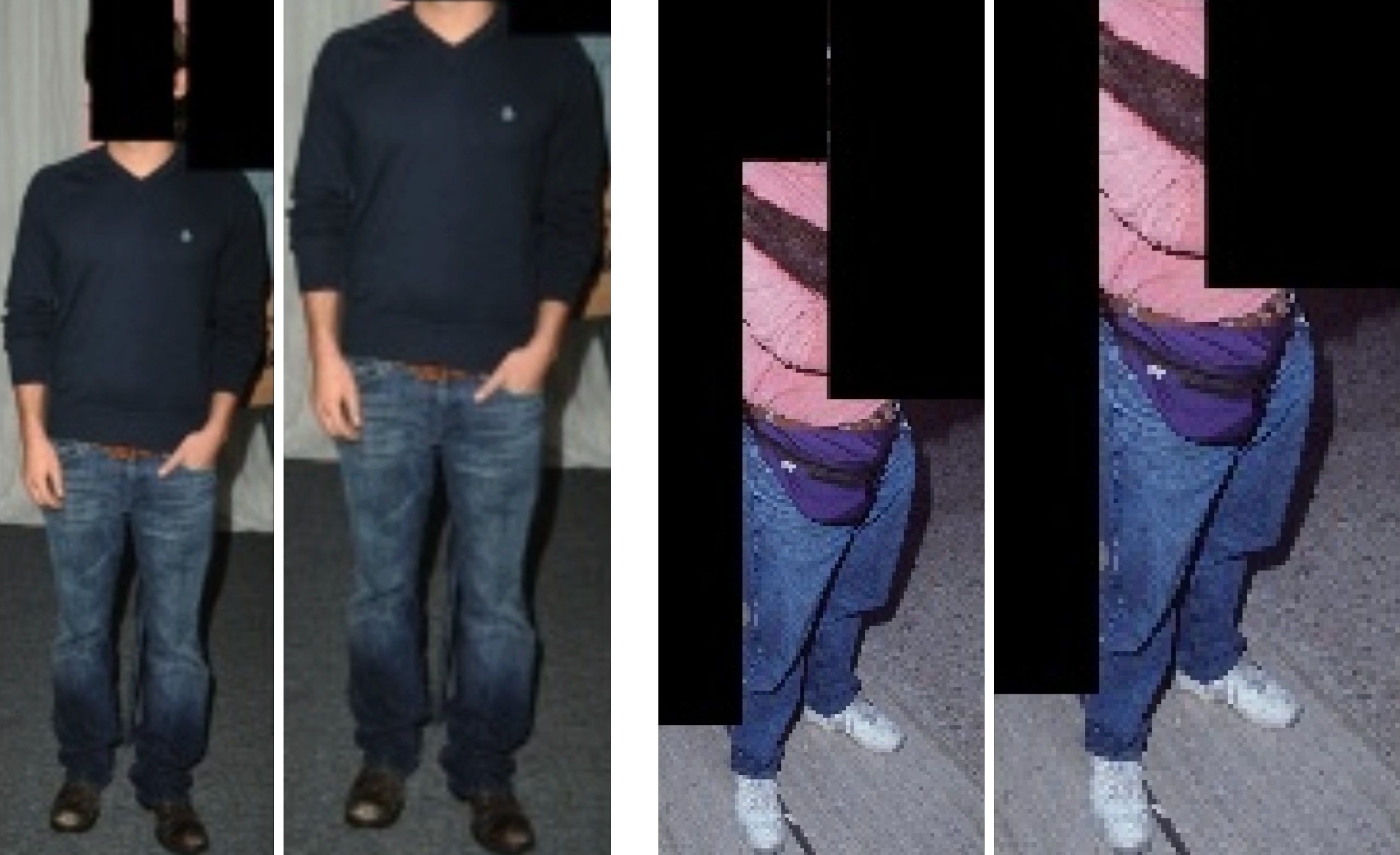}\\
\caption{Body preprocess operation visualization, left is after $DetachObject(b_{i})$ and before $Trim(b_{i})$, right is after.}
\label{fig:cropped_bodies} 
\end{figure}

Unlike faces, body images of persons pose several specific challenges. The body can be heavily occluded, and there can be many different small parts of the body appearing in the image that are not useful for recognition. Such images require a more complex preprocessing approach. Additionally, the nature of bounding boxes introduces another challenge. While face crops rarely contain other people's faces or bodies, body crops often do.
We implemented additional preprocessing steps for body images:
\begin{enumerate}
    \vspace{-0.2cm}\item Check for intersections of the current body bounding box $body_i$ with all detected objects in the image. If any intersection exists, regardless of its size, apply $DetachObject(body_{i})$ that removes all the objects intersected with the $i$-th. This also applies to the paired face crop.
    \vspace{-0.2cm}\item The remaining image may contain unwanted artifacts. To handle these artifacts, we added a trimming operation $Trim(b_i)$. In Figure \ref{fig:cropped_bodies}, result of this operation can be observed. 
    \vspace{-0.2cm}\item If the resulting body image is too small in terms of pixels or size compared to the original crop, it is considered useless and discarded.
\end{enumerate}

\subsection{Performance} \label{section:performance}

\begin{figure}[t]
\centering
\includegraphics[width=8.2cm]{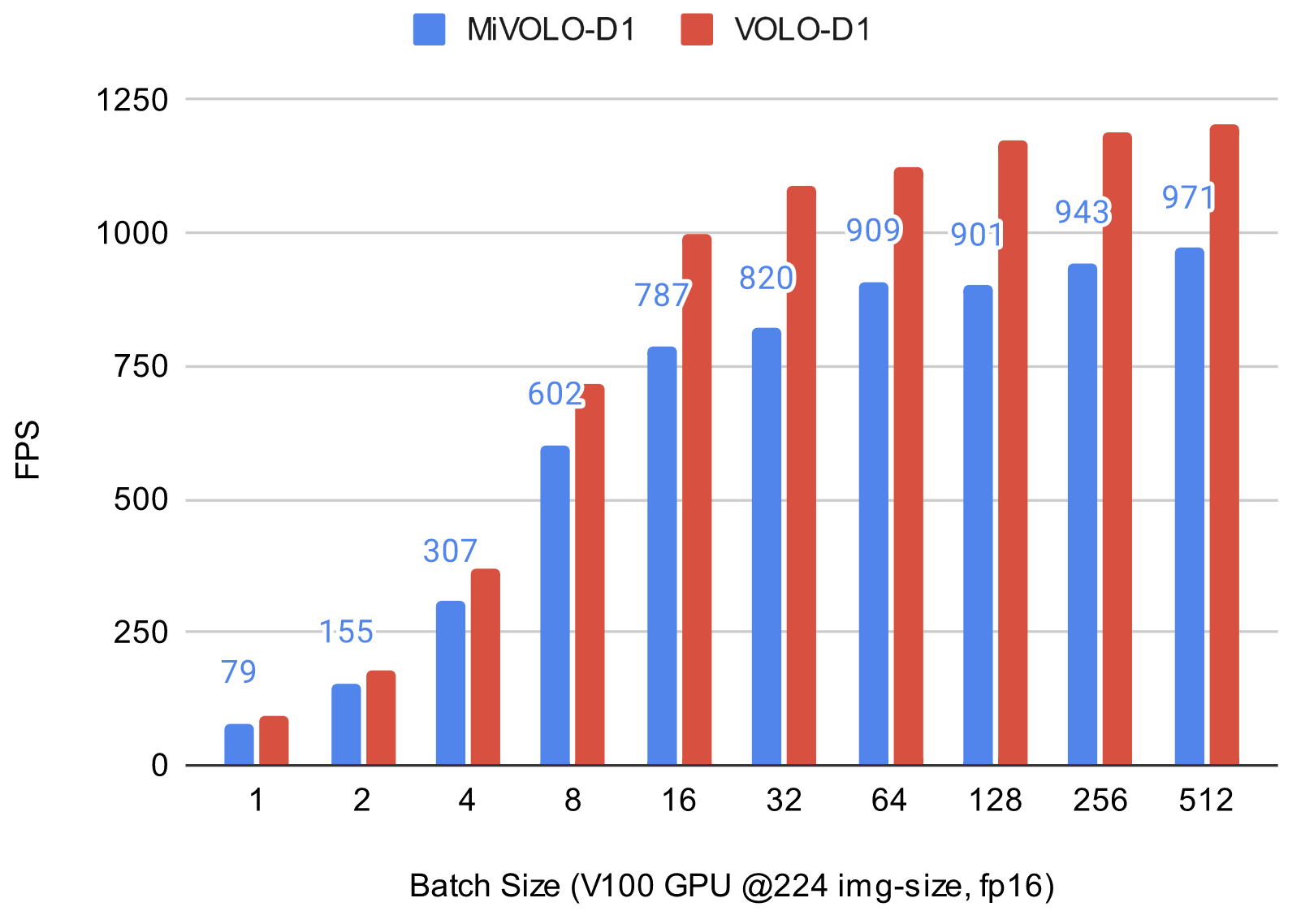}\\
\caption{Model performance comparison with a single V100 GPU: \ModelName-D1 for age and gender estimation by face and body images and single-input VOLO-D1 for age estimation only by face images.}
\label{fig:fps} 
\end{figure}

We consider the VOLO-D1 model variation as our baseline, which consists of 25.8M parameters. In comparison, the \ModelName-D1 model has 27.4M parameters. Figure \ref{fig:fps} demonstrates that while \ModelName-D1 is slightly slower than the original version, it still exhibits high performance. All measurements were conducted using a single V100 GPU with $float16$ precision. When dealing with a single input (even in a mixed batch), we have the option to skip the first $PatchEmbedding$ step for the missing input, leading to a significantly faster inference time.

\begin{table*}[t]
\centering
\begin{tabular}{|lc|c|c|c|c|c|}
\hline
 \multicolumn{1}{|c}{Model} & \multicolumn{1}{c|}{Output} & \multicolumn{1}{c|}{Train Dataset} & \multicolumn{1}{c|}{Test Dataset} & \multicolumn{1}{c|}{Age MAE} & \multicolumn{1}{c|}{Age CS@5} & \multicolumn{1}{c|}{Gender Acc} \\ [1ex] 
\hline
FP-Age \cite{imdbwikiclean} & age & IMDB-clean & IMDB-clean & 4.68 & 63.78 & - \\[0.5ex] 
\hline
VOLO-D1 face$\dag$ & age & IMDB-clean & IMDB-clean & 4.29 & 67.71 & - \\
 &  &  & UTKFace & 5.28 & 56.79 & - \\
 &  &  & Lagenda test & 5.46 & 57.90 & - \\[0.5ex] 
\hline
VOLO-D1 face$\dag$ & age \& gender & IMDB-clean & \textbf{IMDB-clean} & \textbf{4.22} & \textbf{68.68} & \textbf{99.38} \\
 &  &  & UTKFace & 5.15 & 56.79 & 97.54 \\
 &  &  & Lagenda test & 5.33 & 59.17 & 90.86 \\[0.5ex] 
\hline
 CORAL \cite{coral} & age & UTKFace & UTKFace & 5.39 & - & - \\[1ex] 
\hline
 Randomized Bins \cite{random_bins} & age & UTKFace & UTKFace & 4.55 & - & - \\[1ex] 
\hline
 MWR \cite{ordinal_regress} & age & UTKFace & UTKFace & 4.37 & - & - \\[1ex] 
\hline
VOLO-D1 face$\dag$ & age & UTKFace & IMDB-clean & 8.59 & 37.96 & - \\
 &  &  & UTKFace & 4.23 & 69.72 & - \\
 &  &  & Lagenda test & 11.16 & 30.51 & - \\[0.5ex] 
\hline
VOLO-D1 face$\dag$ & age \& gender & UTKFace & IMDB-clean & 8.06 & 41.72 & 97.05 \\
 &  &  & \textbf{UTKFace} & \textbf{4.23} & \textbf{69.78} & \textbf{97.69} \\
 &  &  & Lagenda test & 11.37 & 30.20 & 83.27 \\[0.5ex] 
\hline
VOLO-D1 face & age & Lagenda train & IMDB-clean & 4.13 & 69.33 & - \\
 &  &  & UTKFace & 3.90 & 72.25 & - \\
 &  &  & Lagenda test & 4.19 & 69.36 & - \\[0.5ex] 
\hline
VOLO-D1 face & age \& gender & Lagenda train & IMDB-clean & \textbf{\underline{4.10}} & \textbf{\underline{69.71}} & \textbf{\underline{99.57}} \\
 &  &  & UTKFace & \textbf{\underline{3.82}} & \textbf{\underline{72.64}} & \textbf{\underline{98.87}} \\
 &  &  & \textbf{Lagenda test} & \textbf{4.11} & \textbf{70.11} & \textbf{96.89} \\[0.5ex] 
\hline
\end{tabular}
\caption{Comparison of accuracy of VOLO-D1 models and previous SOTA results. \textbf{Bold} indicates the best model, trained and evaluated on the same datasets. \textbf{\underline{Bold}} indicates the best model with additional train data. \dag\ marks the models that we release to the public domain. }
\label{table:baseline_age_gender_results}
\end{table*}

\section{Experiments} \label{section:experiments}

Our code is based on PyTorch \cite{pytorch} and timm \cite{timm}. We use the VOLO \cite{volo} model as our baseline.

\subsection{Evaluation metrics} \label{section:metrics}

In this section, we present the model's performance using various metrics. For gender prediction and age prediction in classification benchmarks, we utilize the classification accuracy metric.

In regression age benchmarks, the model's performance is evaluated based on two metrics: Mean Absolute Error (MAE) and Cumulative Score (CS). MAE is calculated by averaging the absolute differences between the predicted ages and the actual age labels in the testing dataset. CS is computed using the following formula:

\[ CS_{l} = \frac{N_{l}}{N} \times 100\% \]

Here, \( N \) represents the total number of testing examples, while \( N_{l} \) denotes the count of examples for which the absolute error between the estimated age and the true age does not exceed \( l \) years.

\subsection{VOLO Experiments on Open Source Datasets}

First, we conducted experiments on IMDB-clean and UTKFace datasets to establish a good baseline and identify model limitations. In this section original images, annotations and data splits were taken.

For the age estimation task, \textbf{our baseline model, VOLO-D1}, was trained using only the face input. We employed the $AdamW$ optimizer with an initial learning rate of \num{1.5e-5} and a weight decay of \num{5e-5}. The model was trained for 220 epochs individually on both the IMDB-clean and UTKFace datasets. The base learning rate batch size was set to 192. At the start of training, we performed a warmup with $lr=\num{1e-6}$ for 25 epochs with gradual increase.

The following data augmentations were applied during training:
\begin{itemize}
    \vspace{-0.2cm}\item RandAugment with a magnitude of 22 and bilinear resizing.
    \vspace{-0.2cm}\item Random bounding box jitter for position and size, both with a magnitude of 0.45.
    \vspace{-0.2cm}\item Reprob with $p=0.5$.
    \vspace{-0.2cm}\item Random horizontal flip with $p=0.5$.
\end{itemize}
Additionally, we incorporated $drop$ and $drop$-$path$ with $p=0.32$.

We performed several experiments, exploring different parameters and loss functions. For age estimation, we tried $WeightedFocalMSE$ loss and $WeightedHuber$ loss, but simple $WeightedMSE$ yielded the best performance.

As shown in Table \ref{table:baseline_age_gender_results} our results are state-of-the-art without any additional data or advanced techniques on IMDB-clean and UTKFace datasets. 

\textbf{For the age \& gender VOLO-D1 model}, we followed the same training process. To address the discrepancy in the magnitudes of loss between age and gender, we weighted the gender loss with $w=\num{3e-2}$. We did not change anything else, including the number of epochs.

By adding a second age output to the model, we expected to observe the same effect as reported in the study \cite{multitask_kendall}, where a single model performs better than multiple separate models, leveraging the benefits of learning two tasks simultaneously. And, indeed, we obtained a significantly better MAE for the age, while also achieving impressive accuracy for gender classification. Please refer to Table \ref{table:baseline_age_gender_results} for the detailed results.

\begin{table*}[t]
\centering
\begin{tabular}{|p{2.1cm}|p{1cm}|
p{1.3cm}|p{0.84cm} p{0.84cm} p{0.84cm} |p{0.84cm} p{0.84cm} p{0.84cm}|p{0.84cm} p{0.84cm} p{0.84cm}|}
\hline
\multicolumn{3}{|c|}{} & \multicolumn{9}{c|}{Tested with} \\ [1ex] 
\multicolumn{3}{|c|}{} & \multicolumn{3}{c|}{Face} & \multicolumn{3}{c|}{Body} & \multicolumn{3}{c|}{Face\&Body} \\ [1ex] 
\hline
\hline
 \multicolumn{1}{|p{2.0cm}|}{Model} & \multicolumn{1}{p{1.1cm}|}{Train Set} & \multicolumn{1}{p{1.3cm}|}{Test Set} & \multicolumn{1}{p{0.83cm}}{MAE} & \multicolumn{1}{p{0.83cm}}{CS@5} & \multicolumn{1}{p{0.86cm}|}{Gender Acc}& \multicolumn{1}{p{0.83cm}}{MAE} & \multicolumn{1}{p{0.83cm}}{CS@5} & \multicolumn{1}{p{0.86cm}|}{Gender Acc} & \multicolumn{1}{p{0.83cm}}{MAE} & \multicolumn{1}{p{0.83cm}}{CS@5} & \multicolumn{1}{p{0.86cm}|}{Gender Acc} \\ [1ex]
\hline
VOLO-D1 & Lagenda & IMDB & 4.10 & 69.71 & \textbf{99.57} & - & - & - & - & - & - \\  
 &  & UTKFace & 3.82 & 72.64 & \textbf{98.87} & - & - & - & - & - & - \\   
 &  & Lagenda & 4.11 & 70.11 & 96.89 & - & - & - & - & - & - \\   
\hline
MiVOLO-D1$\dag$ & IMDB & IMDB & 4.35 & 67.18 & 99.39 & 6.87 & 46.32 & 96.48 & 4.24 & 68.32 & 99.46 \\   
 &  & UTKFace & 5.12 & 59.10 & 97.66 & 6.36 & 47.74 & 95.57 & 5.10 & 97.72 & 59.46 \\   
 &  & Lagenda & 5.40 & 58.67 & 91.06 & 10.52 & 31.70 & 87.71 & 5.33 & 59.20 & 91.91 \\ 
\hline
MiVOLO-D1 & Lagenda & IMDB & 4.15 & 69.20 & 99.52 & 6.66 & 47.53 & 96.74 & \textbf{4.09} & \textbf{69.72} & 99.55 \\   
 &  & UTKFace & 3.86 & 72.06 & 98.81 & 4.62 & 63.81 & 98.69 & \textbf{3.70} & \textbf{74.16} & 98.84 \\  
 &  & Lagenda & 4.09 & 70.23 & 96.72 & 7.41 & 49.64 & 93.57 & \textbf{3.99} & \textbf{71.27} & \textbf{97.36} \\
\hline
\end{tabular}
\caption{Comparison of multi-input \ModelName-D1 and single-input VOLO-D1 age \& gender models accuracy. \textbf{Bold} indicates the best model for each benchmark. \dag\ marks the model that we release to the public domain. }
\label{table:mivolo_results}
\end{table*}

\subsection{MiVOLO Experiments on Open Source Datasets}

We made some minor adjustments to the training process for the \ModelName\ model. To reduce training time, we initialized the model from a single-input multi-output VOLO checkpoint. We initialized weights of the $body\ PatchEmbedding$ block with the same weights as the $face\ PatchEmbedding$ block.  The $Feature Enhancer Module$ was initialized with random parameters.

During training, we froze the $face\ PatchEmbedding$ block since it was already trained. We trained the model for an additional 400 epochs, incorporating random dropout of the body input with a probability of 0.1, and random dropout of the face input with a probability of 0.5. Face inputs were only dropped for samples with suitable body crops. If a face input was dropped, the model received an empty (zero tensor) input for $face\ PatchEmbedding$, and the same for empty body inputs.

These techniques were implemented to adapt the model for various mixed cases and to improve its understanding of input images, resulting in enhanced generalization. We also set the learning rate to \num{1e-5}. To preserve the structural integrity of the data, all augmentations, excluding jitter, are applied simultaneously.

The remaining parts of the training procedure are unchanged.

We conducted experiments on the IMDB-clean dataset using our \ModelName. Table \ref{table:mivolo_results} shows a comparison between the single-input VOLO and the multi-input MiVOLO. The results indicate that the best performance across all benchmarks is achieved by using both face and body crops. The model trained on our dataset consistently outperforms the one trained on IMDB.

To evaluate the quantitative performance of the \ModelName\ when only body images are available, we conducted an experiment where all faces were removed from the data. Additionally, we excluded any images that did not meet our specified requirements mentioned in Section \ref{section:multi_input_data_prep}. For IMDB-clean, UTKFace and Lagenda test datasets retained 84\%, 99.6\% and 89\% of images, respectively. Results are displayed in the Table \ref{table:mivolo_results} and Figure \ref{fig:lagenda_mae} (b).

\subsection{\DatasetNameShort \ experiments}

We repetead all previous experiments on our \DatasetNameShort\ trainset. We trained three variants of the model: VOLO-D1 face-only age, VOLO-D1 face-only age \& gender, and MiVOLO-D1 face + persons age \& gender. We kept all training parameters unchanged, following the same configuration as for the IMDB-clean dataset.

Please refer to Table \ref{table:baseline_age_gender_results} and Table \ref{table:mivolo_results} for the results. As expected, the amount of data played a crucial role in the performance of our \ModelName. We observed significant improvements and achieved SOTA results for the \DatasetNameShort, UTKFace, and IMDB-clean datasets by utilizing the face \& body multi-input approach. Remarkably, we also obtained satisfactory results for body-only inference. 

\begin{figure}[t]
\centering
\includegraphics[width=8.2cm]{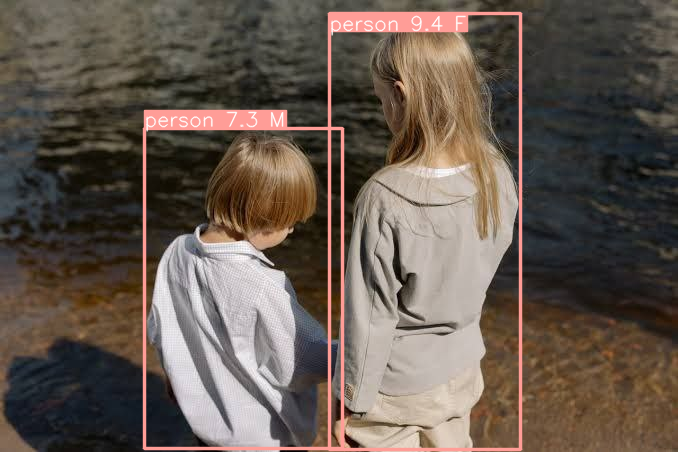}\\
\caption{An illustration of a case where the work is performed without faces on a random picture obtained from the internet.}
\label{fig:boy_girl_example} 
\end{figure}

In Figure \ref{fig:boy_girl_example}, we provide an illustration of a successful recognition result without visible faces in a random picture sourced from the internet.
Model generalizes very well, even though it has never seen images like this with persons shown from the back.

\begin{figure}[htp]
\centering
    \subfloat[face \& body]{%
        \includegraphics[width=7cm]{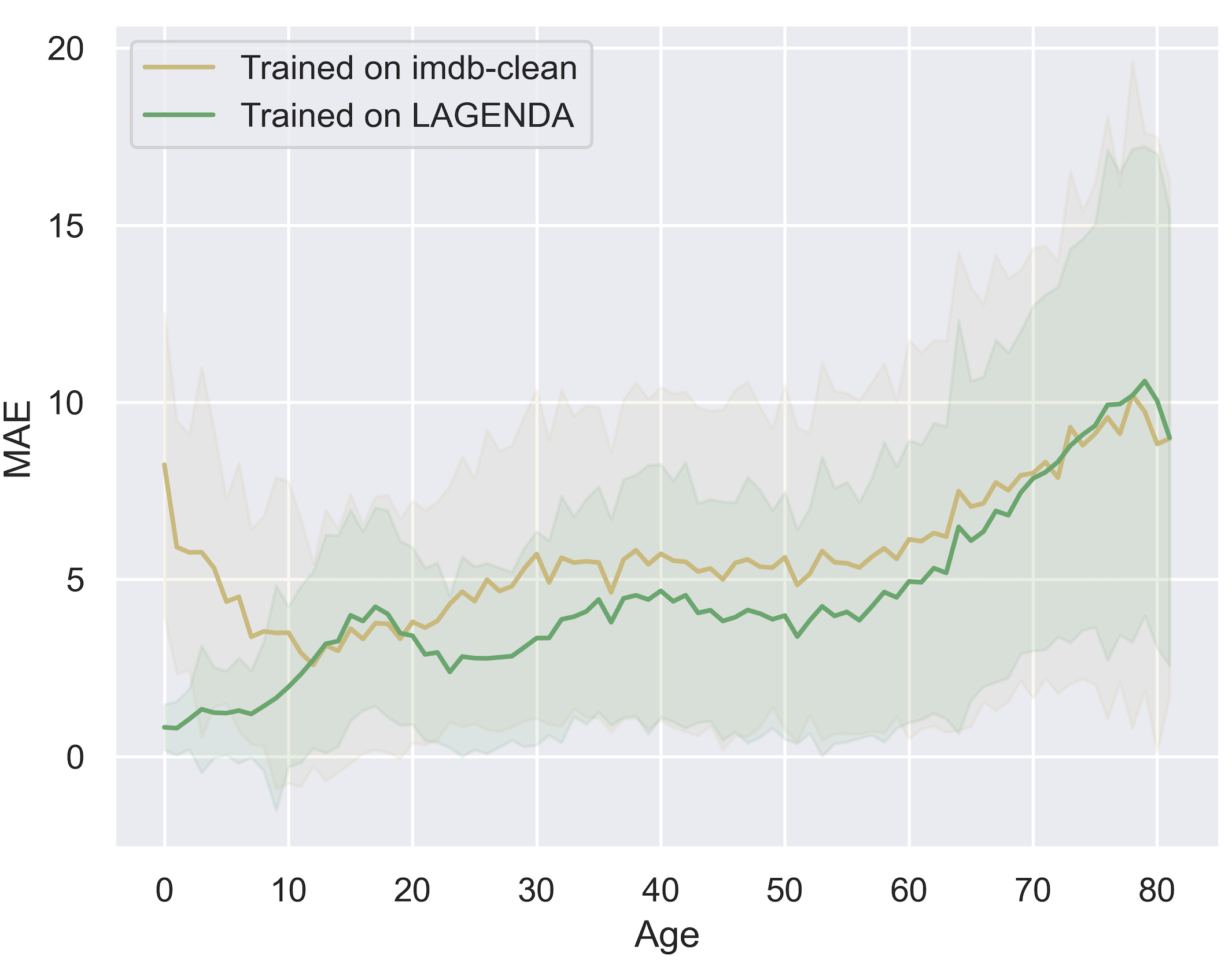}%
    }
    \hfill
    \subfloat[body only]{%
        \centering
        \includegraphics[width=7cm]{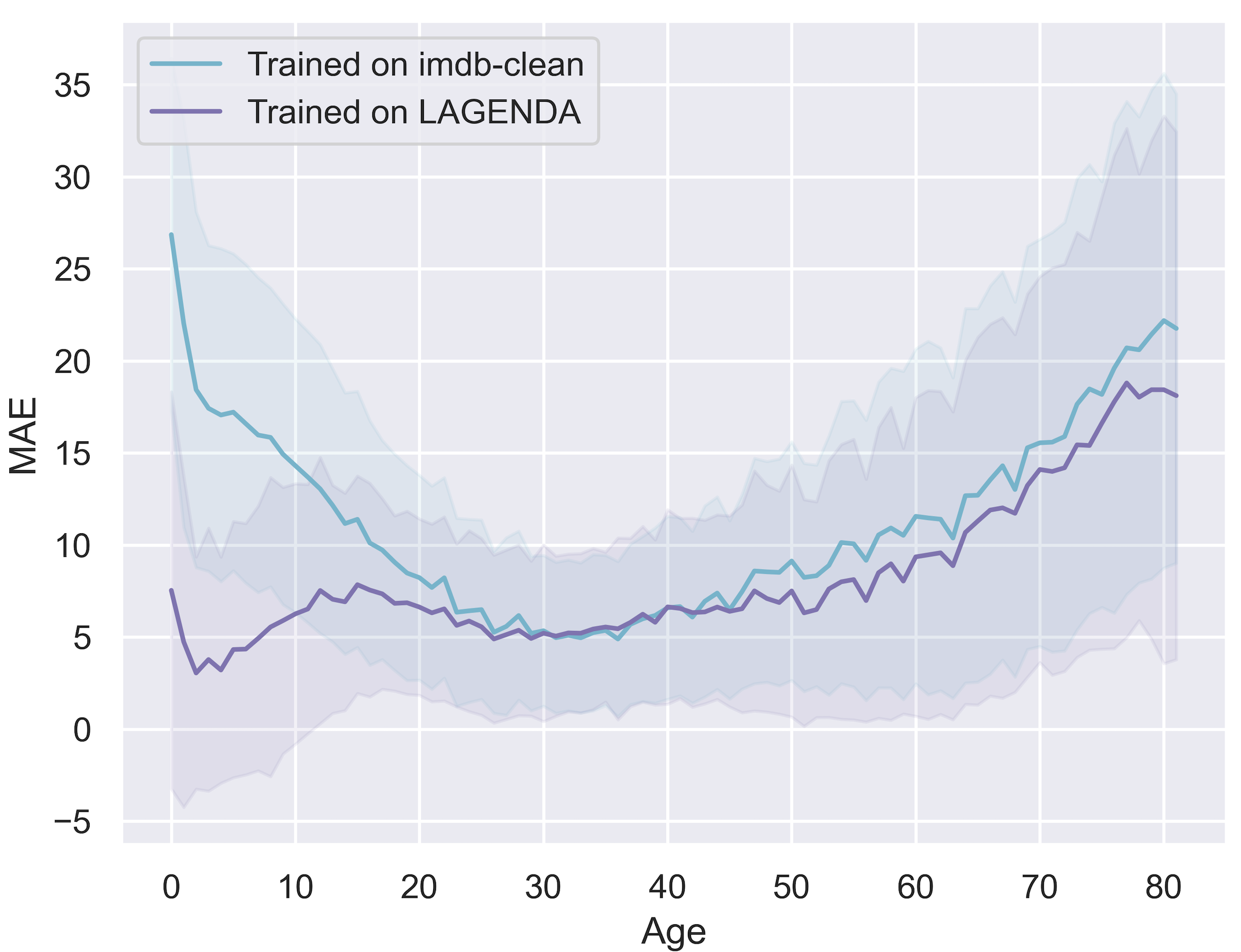}%
    } 
    \caption{Relationship between MAE and age for \ModelName. Tested on \DatasetNameShort\ benchmark using: a) face \& body; b) only body. }\label{fig:lagenda_mae}
\end{figure}

Relationship between MAE and age for final models is shown in Figure \ref{fig:lagenda_mae} (a) and (b).

\subsection{Adience, FairFace, AgeDb benchmarks}

Due to the model's impressive generalization capabilities, we decided to apply \ModelName \ to the AgeDb \cite{moschoglou2017agedb} regression benchmark and to popular classification benchmarks such as FairFace \cite{fairface} and Adience \cite{adience}. As our model was not explicitly trained for classification tasks, we applied our final \ModelName-D1 age \& gender model to FairFace and Adience without any modifications. The only change made was mapping the regression output to classification ranges. As shown in Table \ref{table:classification_results}, we achieved SOTA results for the mentioned datasets without any additional changes.

\begin{table}[h!]
\centering
\begin{tabular}{|p{2.5cm}|p{0.83cm}|p{0.83cm}|p{0.83cm}|p{0.83cm}|} 
 \hline
 \multicolumn{1}{|c|}{Method} & \multicolumn{1}{c|}{Test Set} & \multicolumn{1}{p{0.83cm}|}{Age Acc} & \multicolumn{1}{p{0.83cm}|}{Age MAE} & \multicolumn{1}{p{0.83cm}|}{Gender Acc} \\ [0.5ex] 
 \hline\hline
 FairFace\cite{fairface} & FairFace & 59.70 & & 94.20 \\
 \textbf{MiVOLO-D1 Face\&Body} & FairFace & \textbf{61.07} & & \textbf{95.73} \\[1ex]
  \hline
 DEX \cite{Rothe-ICCVW-2015} \cite{moschoglou2017agedb} & AgeDB & & 13.1 & - \\
 \textbf{MiVOLO-D1 Face} & AgeDB & & \textbf{5.55} & \textbf{98.3} \\[1ex]
  \hline
 MWR \cite{ordinal_regress} & Adience & 62.60 & & - \\
 AL-ResNets-34 \cite{lstm_age} & Adience & 67.47 & & - \\
 Compacting \cite{compacting} & Adience & - & & 89.66 \\
 Gen MLP \cite{retina_arc} & Adience & - & & 90.66 \\
 \textbf{MiVOLO-D1 Face} & Adience & \textbf{68.69} & & \textbf{96.51} \\[1ex] 
 \hline
\end{tabular}
\caption{FairFace, Adience, AgeDB validation results using MiVOLO-D1 trained on LAGENDA train set.}
\label{table:classification_results}
\end{table}

\section{Human level estimation and votes ensembling for age recognition} \label{section:human_level}

\subsection{Human level for age estimation} 

As described in Section \ref{section:datasets}, during the annotation of the \DatasetNameShort, control tasks (honeypots) were generated from the IMDB-clean dataset. A total of 3,000 random examples were sampled for this purpose. Users were not aware of which examples were honeypots and annotated them alongside other tasks. This approach provided a reliable source for estimating the human level of performance in the task.

\begin{figure}[t]
\centering
\includegraphics[width=8.2cm]{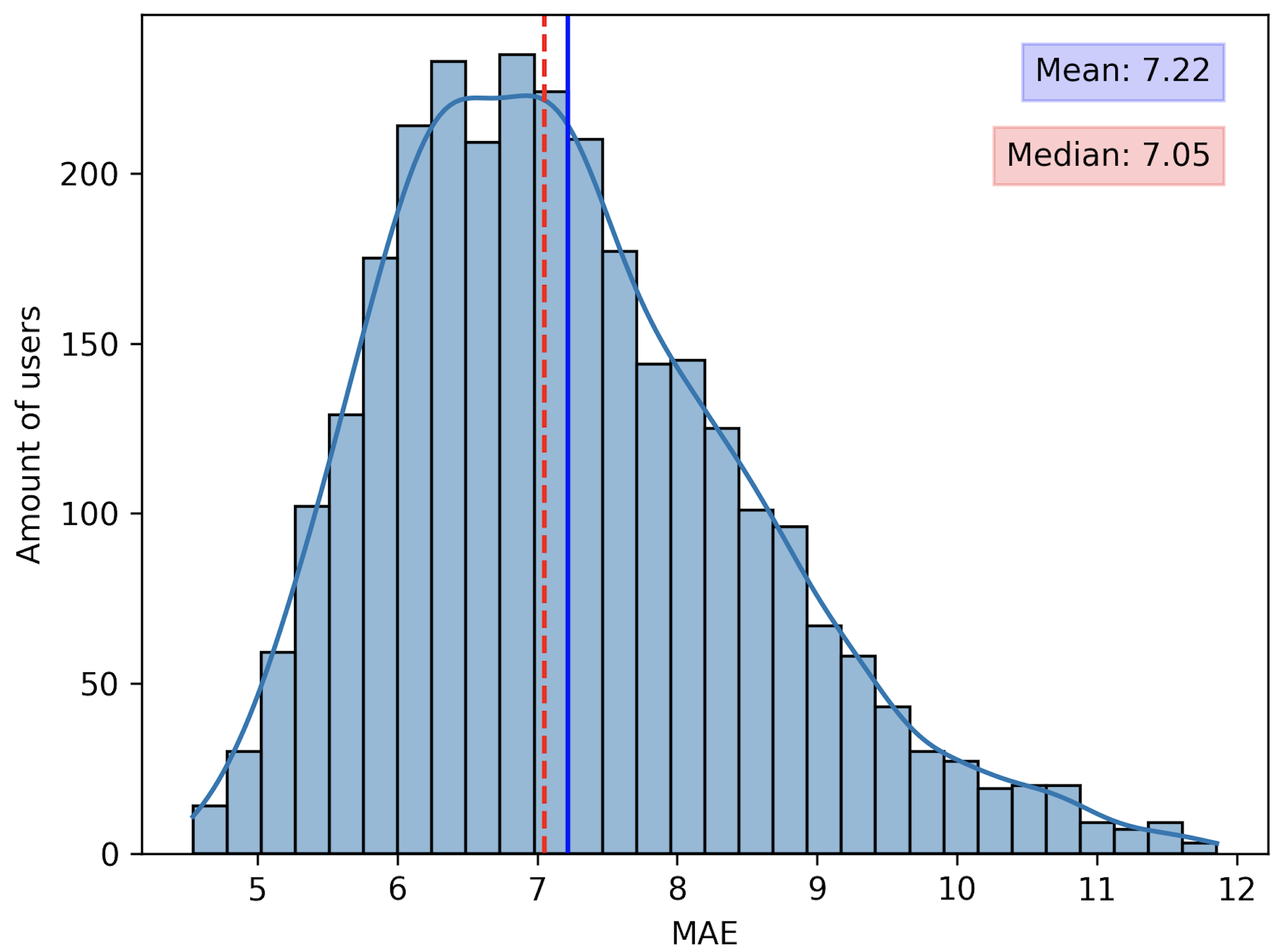}\\
\caption{Histogram of MAE across users, measured on control tasks (n $\ge20$).}\label{fig:human_level}
\end{figure}

Figure \ref{fig:human_level} illustrates the distribution of MAE values among the users. The mean of this distribution is 7.22, and the median is 7.05. The averaged maximum error is 28.56, while the minimum mean error for a specific user is 4.54.

We have briefly described paper \cite{human_vs_machine} in section \ref{section:related_work}. 
We disagree with the method of excluding certain age ranges as it can potentially lead to incorrect conclusions. The authors claimed that their model's accuracy is either equal to or surpasses human accuracy. However, since we can only consider the results obtained on the FG-NET dataset due to the aforementioned issue, we have only one estimation where the model achieved an MAE of 4.6 compared to 4.7 in humans. Given this small difference and the sample size of 1,002 images, the statistical evidence does not appear to be substantial. Furthermore, it is important to note that both datasets have specific visual domains, which can further affect the generalizability of the results. \\
To accurately compare human and machine performance, it is crucial to take into account the entire range of ages and images from the wild domain.

\begin{figure}[t]
\centering
\includegraphics[width=8.2cm]{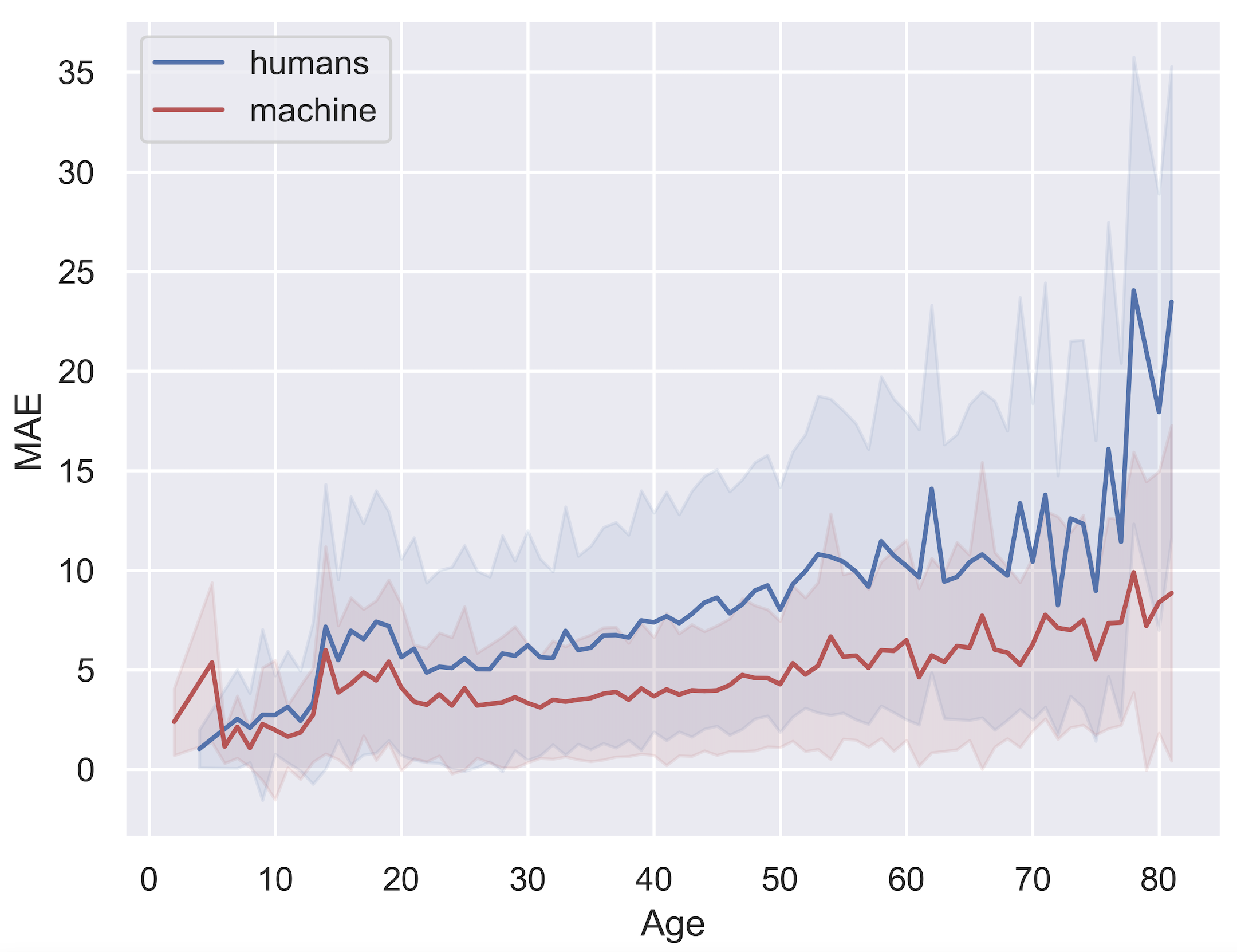}\\
\caption{Scatter plot depicting the relationship between MAE and age on the IMDB-clean dataset, annotated by both human annotators and best \ModelName.}\label{fig:human_error_to_age} 
\end{figure}

As can be seen in Figure \ref{fig:human_error_to_age}, the previous suggestion about low neural network and high human performance in the age range of $[0, 15]$ years no longer holds. 
It turned out that both humans and neural network exhibit an increase in error and its dispersion with the age of the person in the image.

Overall, we can confidently state that our model surpasses human annotators across the majority of age ranges. Furthermore, as shown in Table \ref{table:mivolo_results}, the model achieved a MAE of 6.66 on IMDB-clean with body-only images. This demonstrates that, on average, our model outperforms humans even when considering body-only mode.

\section{Conclusions}
We have introduced a simple yet highly efficient model, \ModelName, that achieves state-of-the-art results on 5 benchmarks, demonstrating its capability to function robustly even in the absence of a face image.

To contribute to the research community, we are providing the weights of the models, which have been trained on Open Sourced data.

In addition, we have enriched and expanded the annotations of 3 prominent benchmarks, IMDB-clean, UTKFace and FairFace. Furthermore, we have developed our own diverse and unbiased \DatasetName, which contains challenging real-world images and is publicly available.

For the task of age annotation aggregation, we employed an intuitive yet remarkably accurate method and evaluated its performance.

Our investigation into the comparison of human and machine accuracy in age recognition tasks revealed that our current model consistently outperforms humans across various age ranges, exhibiting superior overall accuracy.
\section{Future Work and Discussion}

Despite the fact that we achieved our goals, some questions remain open. We still cannot be sure about the physically possible MAE on these or any other age recognition task in computer vision. 

However, the weighted mean from human annotators gives us a very interesting estimation of a certain achievable level in the age recognition task, which is 3.5.

Our approach can be significantly improved by incorporating new class-agnostic segmentation approaches, such as the Segment Anything Model \cite{sam}. These approaches can provide accurate masks for the body, which would be highly beneficial. 

Certainly, even in our very well-balanced dataset, there is a lack of data in the higher age ranges, particularly around 80 years and beyond. As we have shown, the largest contribution to the achieved MAE comes from this range, so it needs to be addressed in future work.

Additionally, this task requires a huge amount of data in order to train a perfect model. However, due to the nature of the task, it is very difficult to obtain it. Therefore, we expect that our method can be combined with Masked Autoencoders \cite{masked_autoencoders} or other scalable self-supervised method.

{\small
\bibliographystyle{ieee_fullname}
\bibliography{egbib}
}

\end{document}